\pdfoutput=1 

\documentclass[letterpaper,twocolumn,10pt]{article}
\usepackage{algorithm}                              
\usepackage{algpseudocode}                          
\usepackage{usenix2019_v3}
\usepackage{amsmath, amsthm, amssymb, mathtools, thmtools}
\usepackage{hyperref, cleveref}
\usepackage{tikz}
\usepackage{caption}
\usepackage{subcaption}
\usepackage[noblocks]{authblk}
\let\oldaffil\affil
\renewcommand{\affil}{\oldaffil[ ]}
\usepackage{soul}
\usepackage{filecontents}
\usepackage{biblatex}
\makeatletter
\newcommand*{\barfix}[2][.175ex]{%
  \mathpalette{\@barfix{#1}}{#2}%
}
\newcommand*{\@barfix}[3]{%
  \vbox{%
    \kern#1\relax
    \hbox{$#2#3\m@th$}%
  }%
}
\makeatother

\begin{document}

\date{}

\title{\Large \bf FedVal: Different good or different bad in federated learning}

\author[1]{Viktor Valadi}
\author[2]{Xinchi Qiu}
\author[2]{Pedro Porto Buarque de Gusmão}
\author[2]{Nicholas D. Lane}
\author[2,3]{Mina Alibeigi}

\renewcommand\Authands{, } 
\renewcommand\Authfont{}
\renewcommand\Affilfont{}

\affil{\textsuperscript{1}AI Sweden\quad \textsuperscript{2}University of Cambridge\quad \textsuperscript{3}Zenseact AB}



\maketitle

\begin{abstract}
Federated learning (FL) systems are susceptible to attacks from malicious actors who might attempt to corrupt the training model through various poisoning attacks. FL also poses new challenges in addressing group bias, such as ensuring fair performance for different demographic groups. Traditional methods used to address such biases require centralized access to the data, which FL systems do not have. In this paper, we present a novel approach \emph{FedVal} for both robustness and fairness that does not require any additional information from clients that could raise privacy concerns and consequently compromise the integrity of the FL system. To this end, we propose an innovative score function based on a server-side validation method that assesses client updates and determines the optimal aggregation balance between locally-trained models. Our research shows that this approach not only provides solid protection against poisoning attacks but can also be used to reduce group bias and subsequently promote fairness while maintaining the system's capability for differential privacy. Extensive experiments on the CIFAR-10, FEMNIST, and PUMS ACSIncome datasets in different configurations demonstrate the effectiveness of our method, resulting in state-of-the-art performances. We have proven robustness in situations where 80\% of participating clients are malicious. Additionally, we have shown a significant increase in accuracy for underrepresented labels from 32\% to 53\%, and increase in recall rate for underrepresented features from 19\% to 50\%.

\end{abstract}
\vspace{-3mm}
\section{Introduction}

Federated Learning (FL) is a novel privacy-preserving machine learning paradigm that collaboratively trains a model across many devices, each using its own local data. As the popularity of machine learning (ML) has exploded in recent years, one of the most significant bottlenecks for ML projects has been the collection of, and access to large, high-quality datasets \cite{roh2019survey}. However, with the growing concerns around privacy and data regulations \cite{voigt2017eu, gdpr}, finding effective ways to gather data while preserving privacy has become increasingly important \cite{roh2019survey}. FL offers a promising approach to address these challenges, and it has been applied in a range of industries, including mobile internet, healthcare, finance, and insurance \cite{li2019abnormal,rieke2020future,dou2021federated}. 

Essentially, FL allows multiple participating devices or systems to collaboratively train an ML model while keeping their data on their own devices rather than gathering it in a single location, helping protect the privacy of the individuals or organizations. In FL, only the model parameters are shared and transmitted back to a server, where they are typically aggregated using a weighted averaging function~\cite{mcmahan2017federated,huang2022learn}. However, this simple approach may not be sufficient in many situations due to the decentralized nature of FL, which introduces challenges such as byzantine failures~\cite{li2020federated2}, data heterogeneity~\cite{li2020federated, karimireddy2020scaffold, qiu2021first, emd}, security and privacy preservation of participants' data
~\cite{kairouz2021advances,bonawitz2016practical,andrew2021differentially}. 

Firstly, machine learning models trained using FL are vulnerable to byzantine failures such as faulty sensors, communication noise, and poisoning attacks \cite{li2021byzantine}. These failures can be caused by compromised clients contributing malicious or faulty global model updates \cite{shejwalkar2022back}. For example, in the context of training an ML model for autonomous cars, faulty sensors, inconsistent connection, or malicious attacks attempting to manipulate the model to gain an advantage could pose a threat. FL models are also vulnerable to attacks on the databases used for training \cite{polap2021meta}, as the central server cannot access the data to verify its integrity. Protecting the ML model against these types of failures is important in FL, particularly because of its decentralized nature, which allows for the inclusion of potentially malicious clients in the training process.

In addition, when working with FL systems, it is typical for the data distributions between clients not to be independent and identically distributed (non-IID), which can pose challenges for the model to effectively and fairly learn from the data \cite{li2022federated,ezzeldin2021fairfed}. Consider the scenario of using FL to train an ML model on data collected from cars. The data distributions gathered by different cars will likely vary significantly due to differences in driving conditions such as location (e.g., rural areas versus cities) and climate. As a direct result of data heterogeneity, the performance of a model trained using FL might degrade when compared to its centralized counter-part~\cite{li2022federated,ezzeldin2021fairfed,li2020federated,karimireddy2020scaffold, qiu2022zerofl}.


Furthermore, one of the main benefits of FL is the ability to train ML models on large amounts of data without having to access or handle sensitive information directly. This is particularly important in the context of data privacy regulations such as the General Data Protection Regulation (GDPR) \cite{gdpr}, which require organizations to protect the personal data of individuals from unauthorized access or use. However, it is important to note that even with federated learning, there is still a risk that sensitive information could be inadvertently revealed or reconstructed from the model's parameters or gradients \cite{bonawitz2016practical,zhu2019deep, zhao2020idlg}. Since federated learning operates by communicating the gradients and the model parameter updates between the central server and the participating clients during each communication round, it is possible to learn sensitive and private information about the underlying data through these gradients and model updates. To address this concern, researchers have developed techniques for ensuring the privacy and security of FL systems, such as differential privacy~\cite{andrew2021differentially} and secure multiparty computation~\cite{bonawitz2016practical}, which make it possible to prove that the sensitive data can not be reconstructed or compromised during the training process. It is important that the methods provided to handle non-IID issues and poisoning attacks are compatible with such privacy-preserving techniques so we can achieve the highest performance while securing access to private data. 
In this paper, we examine the overarching threat model posed by various issues, focusing on realistic Federated Learning settings that address privacy concerns and account for heterogeneity issues. Our primary emphasis is on mitigating large-scale byzantine failures and poisoning attacks, which can have significant consequences.

We examine state-of-the-art solutions to these open problems in federated learning. We also evaluate the effectiveness of these solutions when applied in combination with differential privacy techniques such as norm clipping and noise injection \cite{andrew2021differentially}. Our objective is to uncover potential avenues for enhancing the interrelated aspects of security, robustness, and fairness in FL systems and present innovative solutions to address these challenges simultaneously. 

$\textbf{Contributions} \hspace{10pt} $  To this end, we propose \emph{FedVal}, a novel adaptive server-side validation solution to address the challenges of both poisoning attacks and data heterogeneity. \emph{FedVal} involves scaling the impact of client updates based on their performance to the extent that they have learned from the data. We argue that this approach is more effective than other existing methods that have been proposed, and we demonstrate its efficacy through comprehensive experiments conducted in different FL settings. Our experiments show that our method can provide strong protection against poisoning attacks, even under conditions where $80\%$ of clients are malicious and in the presence of differential privacy techniques. Additionally, \emph{FedVal} is able to handle the fairness challenges posed by non-IID data effectively, for example by completely salvaging quantity distribution skew situations where other solutions struggle. We have also demonstrated that \emph{FedVal} exhibits computational efficiency with the potential to deliver substantial robustness using only $10$ validation elements per class.



Overall, our results highlight the importance of considering the interplay between poisoning attacks, data heterogeneity, and privacy-preserving techniques in the design of FL systems. We believe our work will inspire further research in this direction and contribute to the development of more robust and secure FL systems.

\vspace{-3mm}
\section{Background \& Related work}

In this Section, we provide an overview of the federated learning paradigm (Section \ref{sec:backfl}), poisoning attacks (Section \ref{sec:backpoi}), heterogeneity issues (Section \ref{sec:backnoniid}), and differential privacy (Section \ref{sec:backdp}). Possible solutions to poisoning attacks are presented in Sections \ref{sec:defense1} and \ref{sec:defense2}, while solutions to data heterogeneity issues are found in Section \ref{sec:heterosolution}.

\subsection{Federated learning}\label{sec:backfl}

\begin{figure}[h]
    \centering
    \includegraphics[width=0.40\linewidth]{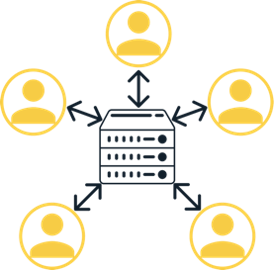}
    \caption{Illustration of FL in centralized setting: model parameters are sent back and forth between a server and clients.}
    \label{fig:fl}
\end{figure}

FL uses a decentralized paradigm that operates over communication rounds, as illustrated in Figure \ref{fig:fl}.  At the beginning of each communication round $t$, a fraction $r$ of $N$ clients is selected by the central server to participate in training. The server sends the current model's parameters $\theta^t_g$ to the selected clients $S_t$, which will then apply a training function $f$ on their local data $D_d$ to update their local models to $\theta^{t+1}_d = f(\theta^t_g, D_d)$. Finally, the updated models $\theta^{t+1}_d$ are sent back to the server, which aggregates them into the next global model $\theta^{t+1}_g$.

There are several common methods to aggregate the local model updates received from clients. The most commonly applied approach is to perform a weighted averaging of the models according to the number of local data samples used for training \cite{beutel2020flower,li2022federated,qiu2021first,qiu2022zerofl}:

\begin{equation}
    \theta^{t+1}_g = \sum_d^{S_t}{\frac{\theta^{t+1}_d|D_d|}{\sum_d|D_d|}}
\end{equation}

Once the global model has been updated, the central server can start a new communication round.

\subsection{Poisoning attacks} \label{sec:backpoi}

FL is vulnerable to poisoning attacks, in which a malicious client attempts to corrupt the global model by returning an altered model. Poisoning attacks can occur in a number of ways, such as data poisoning or model poisoning \cite{shejwalkar2022back}. In addition, these attacks can be classified as targeted or untargeted, depending on whether the adversary aims to compromise the model on a specific function, such as a label or writing style, or seeks to cause general harm.

Regarding poisoning attacks in FL, it is common for a subset of clients to be compromised and collaborate in an attempt to corrupt the global model while avoiding detection. For example, if an adversary has full access to both the model and the data between training rounds, they will likely perform a model poisoning attack. These attacks are particularly potent as they enable the attacker to manipulate the model in a way that causes maximum damage after aggregation while still evading detection~\cite{shejwalkar2022back, shejwalkar2021manipulating}. In contrast, if the adversary only has access to the data, such as by breaching a database, this limits the type of attacks they can perform. In most research, data poisoning attacks are usually made by label flipping~\cite{li2019abnormal,panda2022sparsefed,sun2019can,li2021byzantine}. Notably, for data poisoning attacks, the percentage of malicious clients is likely to be higher as it is easier to only gain access to the data.

In addition to the distinction between data and model poisoning, attacks can also be targeted or untargeted. Targeted attacks aim to compromise a specific function of the model, while untargeted attacks aim to harm the model in any possible way~\cite{shejwalkar2022back}. Research has demonstrated that even a single compromised client is sufficient to completely compromise the global model if no protection is in place~\cite{fang2020local, shejwalkar2022back, yin2018byzantine}. As such, it is crucial to develop techniques for detecting and defending against poisoning attacks in FL systems to ensure the integrity and robustness of the global model.

In this work, we employ two state-of-the-art poisoning attacks based on the methods proposed by Sun et al. \cite{sun2019can} and Shejwalkar et al.~\cite{shejwalkar2022back}. Sun et al. introduced a backdoor attack using label-swapping techniques, in which attackers aim to cause the model to misclassify a specific label with distinct features as another label. Shejwalkar et al. proposed a data-dependent model poisoning attack using projected gradient ascent to manipulate weights in the most harmful direction. Their attack leverages knowledge of benign gradients to specifically target the multi-Krum defense. Our approach is a more generalized variant of this type of attack, which instead statically scales the norms of the attacking clients by a factor that inflicts the most damage to the model.

\subsection{Data Statistical Heterogeneity} \label{sec:backnoniid}

Statistical heterogeneity in federated learning refers to the fact that individual clients might have different data distributions, which can make decentralized training more challenging. This work focuses on one of the most challenging tasks derived from data heterogeneity, i.e., ensuring group fairness. 
Group fairness refers to not biasing the global model toward any specific demographic group having a common data distribution, such as race or gender \cite{ezzeldin2021fairfed}. Traditional ML methods for handling fairness, such as re-sampling \cite{caton2020fairness} and re-weighting \cite{krasanakis2018adaptive}, require the data to be centralized, which is not possible in a federated setting \cite{kairouz2021advances}. Therefore, new approaches must be developed to ensure that the model is fair and unbiased concerning different groups of users, such as those defined by demographic characteristics or geographic locations.

\subsection{Differential privacy} \label{sec:backdp}

Recent studies in the field of ML have demonstrated the potential for extracting information or reconstructing the training data from the model updates in FL \cite{andrew2021differentially,triastcyn2019federated, bonawitz2016practical}. This presents a significant challenge when working with sensitive data that is intended to be private. One of the key use cases for FL is the ability to analyze sensitive data without violating data privacy regulations. To ensure the security of these private data, it is crucial to implement mechanisms to protect it from unauthorized access. One way to achieve this is by incorporating differential privacy (DP) techniques into the FL framework. These methods typically involve working with the norms $\lVert \Delta \theta^{t+1} \rVert = \lVert \theta^t - \theta^{t+1} \rVert$ of client updates, by methods such as clipping and adding noise \cite{andrew2021differentially, wei2020federated}, to prevent the extraction of important information from the clients. There is always a trade-off between model performance and model security, and carefully tuning the hyper-parameter of DP is required to ensure that the training model is effectively protected while maintaining a reasonable model performance. To find this tuning is a topic under ongoing research \cite{kairouz2021advances}. 

\subsection{Defenses preliminary} \label{sec:defense1}

Defenses against poisoning attacks in FL follow a common theme of removing a set number of outlier weights or outlier clients \cite{shejwalkar2021manipulating,shejwalkar2022back,panda2022sparsefed,blanchard2017machine,fang2020local,yin2018byzantine,li2020learning,li2019abnormal}. We make the argument that this approach is not feasible since the number of malicious clients cannot be known and will likely lead to issues where vital information is lost. Methods for estimating a possible percentage of malicious clients can be found in literature \cite{panda2022sparsefed,shejwalkar2022back}, but these are limited to specific cases and provide no tangible guarantees.

In Section \ref{sec:expmal}, we conducted a mathematical investigation of how high the ceiling is for a potential number of malicious clients present in a round for certain situations. We conclude that even with the knowledge of how many malicious actors might be present, we must still ensure robustness for many more clients than there are present due to the randomness of client selection. By this argument and the argument that the number of malicious clients can not be known, we assume that the threshold for how many potentially malicious clients or weights must be removed will need to be a relatively high percentage. Firstly, because we do not know the number of potentially malicious clients, and secondly, because the threshold must be quite a bit higher than the potential number of malicious clients.

The main problem with this is that using specific criteria to remove a high number of clients each round will cause issues. For example, this can lead to underrepresented clients with unique data never being allowed participation in the making of the model, which could potentially lead to a less accurate and fair model. Therefore, it is crucial to find a more effective approach to defending against poisoning attacks in FL.

\subsection{Existing defenses} \label{sec:defense2}

In FL, several defenses have been proposed to protect against poisoning attacks. These defenses can be broadly grouped into three categories: loss function-based rejection defenses, outlier-based defenses, and norm-bounding defenses.

Loss Function-based Rejection (LFR) \cite{fang2020local} is one such defense, which validates each client by using a server-side validation dataset. Clients that have the most amount of loss are removed from the aggregation process. LFR has been shown to be one of the stronger defenses against poisoning attacks, however, it suffers from the issue of needing prior knowledge of the potential number of malicious clients. This can be a significant drawback as mentioned earlier in this section. It can also be noted that extensively validating each client update may be computationally heavy.

Another group of defenses are outlier-based defenses, some euclidean-distance-based outlier detection defenses are multi-Krum \cite{blanchard2017machine} and trimmed mean \cite{yin2018byzantine}. These defenses remove a set number of outliers either in weights or updates. The removal is based on the euclidean distance between all client updates norms. While, in theory, these approaches look promising, they run into problems when the data is non-IID and has issues against model poisoning attacks that are scaled to make the malicious client as close as possible to being flagged as an outlier. A recent review has shown that the theoretically guaranteed protection claimed by these papers does not hold under state-of-the-art model poisoning attacks and the model is vulnerable with only 10\% malicious clients \cite{shejwalkar2022back}. In addition, these defenses would likely be even less effective in FL settings that use differential privacy, as adding noise to the weights would increase the variance between benign clients even more. This would make it more difficult to identify outliers and increases the chances of mistakenly identifying benign clients as malicious.

Researchers have also proposed cosine-similarity-based defenses against poisoning attacks, such as with FLTrust \cite{cao2020fltrust} and FoolsGold \cite{fung2018mitigating}, which do not assume knowledge of the number of malicious clients. FLTrust proposes a defense that shares many similarities with the one we propose in our work; similar to our approach, they utilize a small server-side dataset, but instead of calculating the loss, they compare the direction of gradients with those observed during server-side training on that dataset. FLTrust also scores each client, but based on similarity instead of loss. They then weigh client updates by that score in the aggregation. Despite the initial promise of these cosine-similarity-based defenses, even under attacks with a large percentage of malicious clients, recent research has demonstrated that they are susceptible to model poisoning attacks that focus on specific key weights within the model, such as the attacks presented in the work of Kasyap et al. \cite{kasyap2022hidden}.

Other researchers have proposed using auto-encoders trained on validation data for anomaly detection as a defense mechanism against adversarial attacks, such as those described in Li et al. \cite{li2019abnormal} and Li et al. \cite{li2020learning}. These defenses also involve removing a set number of outliers, which likely will cause issues as discussed previously. We also believe that these defenses are susceptible to state-of-the-art model poisoning attacks, similar to other anomaly detection methods.

Another category of outlier-based defenses capitalizes on singular-value decomposition techniques, exemplified by the approach presented by Shejwalkar et al. \cite{shejwalkar2021manipulating}. Despite demonstrating potential, a primary disadvantage of singular-value decomposition resides in its cubic time complexity, which constrains the complete examination of the model. As a result, investigators often resort to tactics such as random sampling of model dimensions, as illustrated by Shejwalkar et al.\cite{shejwalkar2021manipulating}. Considering the evolution of new attacks capable of significantly impairing model performance by merely targeting a few weights \cite{kasyap2022hidden}, we argue that singular-value decomposition defenses, like the one advanced by Shejwalkar et al. \cite{shejwalkar2021manipulating}, may be nearing obsolescence.

Norm-bounding \cite{sun2019can} is another popular defense mechanism, which involves binding the norms for weights in client updates. The theory behind this defense is that for an attack to be successful, the attacker would need to have larger norms to move the model into an undesirable position. Binding the norms is a widespread practice that is often used in differential privacy \cite{andrew2021differentially} and has also been adopted by more recent poisoning defenses such as SparseFed \cite{panda2022sparsefed}, another type of outlier-based defense, that argues that all defenses should also add norm bounding as a measure to strengthen their defense. While binding the norms is effective at reducing the impact of a poisoning attack, the impact of an attack on a system that has only norm-binding as the defense measure is still relatively large \cite{shejwalkar2022back,sun2019can}. Additionally, we believe SparseFed might be susceptible to state-of-the-art model poisoning attacks, similar to other anomaly detection methods.

In conclusion, while several defenses have been proposed to protect against poisoning attacks in FL, each has its weaknesses and limitations. Outlier-based defenses may struggle against state-of-the-art model poisoning attacks, and norm-bounding defenses may not provide a strong enough defense. On the other hand, previous loss function-based defenses, while providing a strong defense, still assume knowledge of the number of malicious clients.

\vspace{-3mm}
\subsection{Heterogeneity solutions} \label{sec:heterosolution}

For issues related to data heterogeneity, a significant amount of research has been conducted on preventing client drift, such as using techniques like SCAFFOLD \cite{karimireddy2020scaffold} and FedProx \cite{li2020federated}. While these methods have proven effective in creating models that are more representative of the majority of clients, they might make it difficult in systems where de-biasing and fairness considerations are necessary, which we have investigated in section \ref{sec:fairexp}.

Handling group fairness, which is commonly caused by uneven underlying data distributions \cite{kairouz2021advances}, is an area in FL that has not been widely explored. Recent works such as FairFed \cite{ezzeldin2021fairfed} and the work of Du et al. \cite{du2021fairness} have begun to address this issue. Prior research on fairness in FL often employs re-weighting schemes similar to the one presented in our work. FairFed \cite{ezzeldin2021fairfed} proposes a solution where clients analyze their data and send back information to the central server for aggregation. However, sending additional information by clients to the central server can create risks for privacy \cite{kairouz2021advances}.


\vspace{-5mm}
\section{Methodology} \label{sec:method}

In this section, we present our innovative method \emph{FedVal}. \emph{FedVal} attaches a weight based on bias reduction and relative performance of each client model to the aggregation of their updates, which allows a more dynamic and nuanced approach to protecting the global model. In Section \ref{sec:motivations} we motivate the advantages of using \emph{FedVal} and in Section \ref{sec:fedval} we delve into the design of \emph{FedVal}.

\vspace{-3mm}
\subsection{Motivations for \emph{FedVal}} \label{sec:motivations}

\emph{FedVal} is an innovative method that aims to protect the global model from malicious client attacks while maintaining the robustness and fairness of FL training. 

One key advantage of score-based methods like \emph{FedVal} is that they never discard good model parameters. Other solutions that commonly remove a set number of clients' updates or norms in each round \cite{shejwalkar2021manipulating,panda2022sparsefed,blanchard2017machine,fang2020local,yin2018byzantine,li2020learning,li2019abnormal} can complicate dealing with data heterogeneity issues since the outliers that are removed each round might contain vital information for the model.

\begin{figure}[H]
\centering
\includegraphics[width=0.9\linewidth]{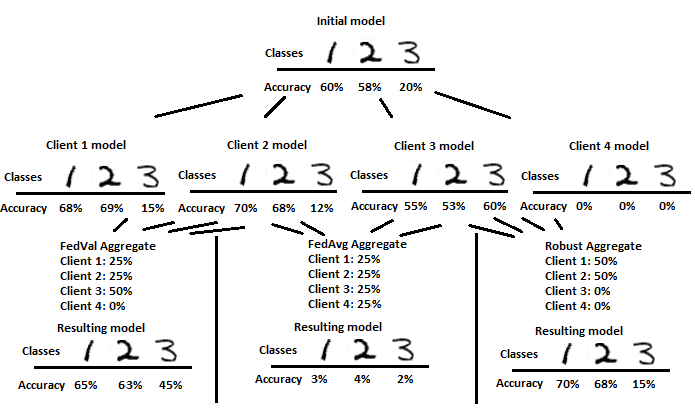}
\caption{Toy example of a federated learning situation with four clients being aggregated, two majority clients with skewed data, one client with information that is missing in the majority of clients, and one malicious client.}
\label{fig:example}
\end{figure}

While the primary goal of \emph{FedVal} is to protect the global model from malicious client updates, we have also incorporated a bias reducer term for further improvement from the average clients' model. The term aims to decrease the potential negative effects caused by the defense mechanism to improve the fairness of the FL system and model performance in the non-IID data distribution situation.

\emph{FedVal} aims to give a secure solution to the interplay between fairness and robustness in FL as illustrated in an example in Figure \ref{fig:example}. In this example, there is a total of $3$ classes in the classification task, and $4$ clients are selected in an FL communication round. $3$ of the clients have updated the model in similar directions, but they are missing one of the three classes. One of the client models is slightly worse than the first two but has information on the class that they are missing. The last client is malicious and has sent back a completely ruined model.  Figure \ref{fig:example} highlights the advantage of \emph{FedVal} as the algorithm that is able to notice the extra contribution from the third client as it has updated the model with new information that other clients are missing, hence giving it higher aggregation weight. \emph{FedVal} can also identify the malicious client and give it zero weight for aggregation.

On the other hand, other robust aggregators would typically favor the two most similar clients and use them for the aggregation. If a standard averaging method such as FedAvg is used, the new model would be completely deteriorated by the malicious update. But even in a situation with no malicious updates, the new information from client $3$ is, in most cases, lost since the norms with the specific information for class $3$ get under-valued compared to those of the majority client update models.

\subsection{FedVal} \label{sec:fedval}

In this section we give a comprehensive explanation of the \emph{FedVal} algorithm, its scoring function, and its various implementation methods. 

The core of \emph{FedVal} is the score function, which is used to score each client based on their performance, as determined by a server-side validation dataset. Let $S(\theta_d^t)$ be the score for client $d$ at round $t$. The main purpose of the score function is to extract relevant features for the model and eliminate unwanted or redundant features. 

We consider a $K$ class classification problem defined over a label space $\mathcal{Y} = [K]$, where $[K] = \{ 1,...,K \}$. Conceptually the score function can be represented as a summation over different labels with a bias reducer term multiplied by a slope term. let $\barfix{\Bar{\pmb{L}}_k}$ be mean validation loss for validation data with label $k$, $\barfix{\Bar{\pmb{L}}_{avg}}$ be the average validation loss over the whole validation set, and $MAD$ mean absolute deviation as explained in Equation \ref{eq:mad}. $div_{k,d}$, loss diversion term, is defined to be the difference between loss on the validation set with only label $k$ using the local model from client $d$ and the average loss from all clients on the validation set with only label $k$. Apart from the dynamic parts, there is a term ($Cs_1$) that gets summed at the end, which gives an average client some baseline score, ensuring that none of the clients' contributions will be unnecessarily missed. We use $C = 3$ in our experiments, but it is worth noting that this hyper-parameter can be improved based on each case. The full formula of the score function can be found below in Equation \ref{eq:ConScore}, this score is later used for aggregation as demonstrated in Equation \ref{eq:fedval}. 

\begin{equation}\label{eq:ConScore}
    S(\theta_d^t) = \sum_{k=1}^{K} (\underbrace{\max(1, \left(\frac{\barfix{\Bar{\pmb{L}}}_{k}}{\pmb{\barfix{\Bar{L}}}_{avg}}\right)^{s_2})}_\text{bias reducer term}*\underbrace{\frac{s_1 *div_{k,d} }{MAD_{k}}}_\text{slope term} + Cs_1)
\end{equation}
\begin{equation}\label{eq:div}
div_{k,d} = \barfix{\Bar{\pmb{L}}}_{k} - \pmb{L}_{k,d}
\end{equation}

In Equation \ref{eq:ConScore}, the score function is summed over the label space. However, it can also be extended with other dimensions that matter, such as overall loss and demographic groups. Note that the hyper-parameters $s_1$ and $s_2$ can be different and thus need to be altered if other dimensions are chosen. Alterations within the label space may also be desirable if fairness for certain important labels is essential.


The slope term is determined by the loss diversion term and the MAD (mean absolute deviation) term. The MAD is defined as Equation \ref{eq:mad}. Assume there are $M$ total validation samples. 
Then, MAD is the loss deviation of each validation sample from the average loss.

\begin{equation}\label{eq:mad}
    MAD = \frac{\sum_{m=1}^{M}{| \pmb{L}_m -\sum_{i=1}^{M} \pmb{L}_i|}}{rN}
\end{equation}

Regarding the bias reducer term in Equation \ref{eq:ConScore}, it divides the average loss on the label $k$ by the average loss on all labels. If the updated model performs worse for label $k$, hence higher average loss $\Bar{\pmb{L}}_k$, the bias reducer term would be above $1$. Thus, the term scales the whole score higher if label $k$ is performing worse than average. Therefore, the bias reducer term can in theory help balance the model to make the global model focus more on the label that is performing worse than other labels. 

\begin{figure}[h]
\centering
\includegraphics[width=0.5\linewidth, trim=1cm 0cm 0cm 0cm]{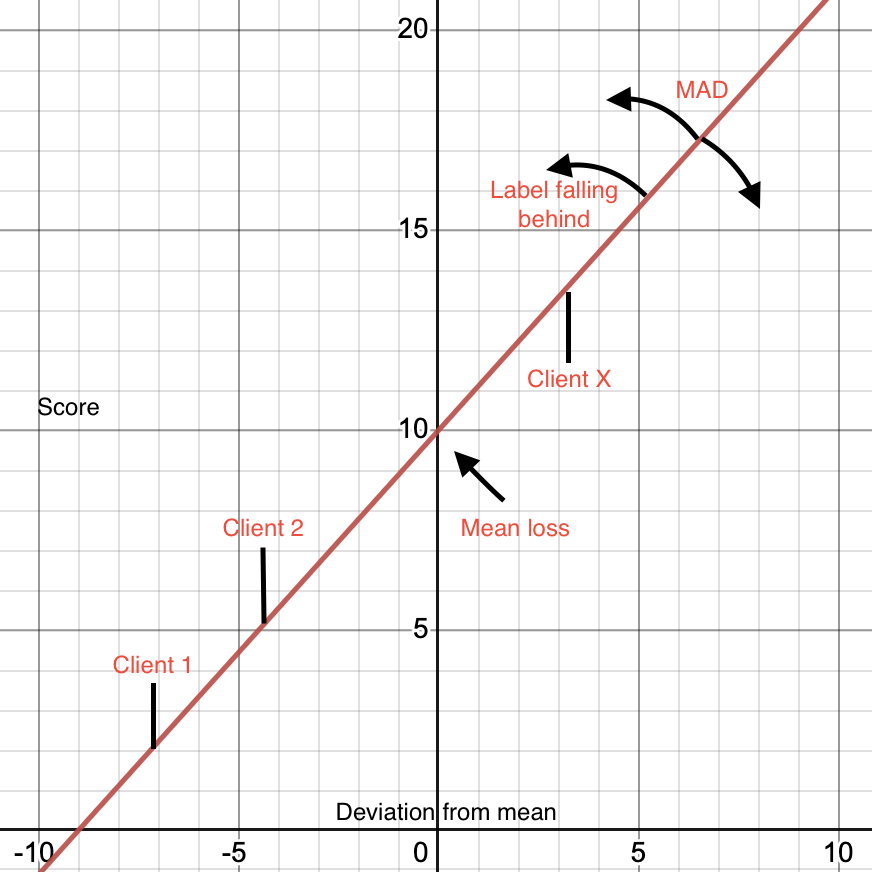}
\caption{Example of how the score is set for a specific label. The slope of the graph gets steeper exponentially by how much this label is behind other labels. The effect of larger or smaller deviations between client loss for the label increases or decreases the slope.}
\label{fig:slope}
\end{figure}

An illustration of how a slope behaves in the \emph{FedVal} equation for a specific label can be seen in Figure \ref{fig:slope}.

The two dynamic terms in Equation \ref{eq:ConScore} is paired with two hyper-parameters $s_1$ and $s_2$. The $s_1$ parameter - which is paired with the slope term - is a booster that steepens the slope. On the other hand, the $s_2$ parameter - which is paired with the bias reducer term - increases how important it is for a dimension to not underperform compared to other model functionality. Something to note is that the bias reducer term has polynomial growth, which makes further underperformances matter more severely. This nuanced approach gives a potential implementer of \emph{FedVal} possibilities to decide which labels or dimensions should be prioritized and promote a more balanced model.

The $s_2$ parameter is designed to be adaptive, which is achieved by evaluating various $s_2$ values on the validation data set and selecting the value that results in the minimum loss. To obtain a fair global model, which performs equally well for each class, we created a balanced validation set representative of the model's end goal. By choosing $s_2$ that achieves the minimum loss, we can push the global model towards a more balanced and fairer performance. In our experiment, we initialize $s_2$ to be $3$, and in each round, check the set $[s_2, s_2+0.5, s_2-0.5, s_2-5, s_2+5]$ for the optimal $s_2$.



The FL training implementing  \emph{FedVal} will follow the standard communication round. At the start of each communication round, a subset of clients is selected, and current global weight parameters will be broadcast to these selected clients. Then, these selected clients will perform local training and return their updated models to the server. The server validates all clients over a validation dataset and scores them as demonstrated in Equation \ref{eq:ConScore}. Lastly, the model parameters will be aggregated using the scores of each client as weight according to Equation \ref{eq:fedval}. The complete \emph{FedVal} algorithm can be seen in Algorithm \ref{algo:fedval} in Appendix \ref{app:algo}.

\begin{equation}\label{eq:fedval}
    \theta^{t+1}_g = \theta^{t}_g + \sum_{d}^{S_t}{\frac{S(\theta_d^{t+1})\Delta \theta^{t+1}_d}{\sum_{d}^{S_t}{S(\theta_d^{t+1})}}}
\end{equation}

The salient dependency of \emph{FedVal} on mean values for client updates is worth noting, as it could theoretically be prone to colluding attacks by multiple malicious clients as outlined in \cite{panda2022sparsefed, shejwalkar2022back}. Under such circumstances, one client could manipulate the mean, enabling other clients to subtly introduce smaller attacks. However, in the classification problem-focused experimental setting of this study, this potential vulnerability is mostly irrelevant due to the inherent limit on classification loss functions. The potential vulnerability of \emph{FedVals} in regression problems remains unexplored, which constitutes an area for future work. A viable approach to this issue might involve constraining the loss function within a reasonable range of values.

Time complexity is a crucial factor when considering defenses that validate clients on server-side validation data. It has been previously noted that exhaustively testing each client on a validation dataset can be computationally intensive \cite{li2020learning}. However, it is important to note that the actual time complexity of validating each client can be reduced by utilizing parallel processing techniques, which results in a time complexity of $O(wv)$ in parallel, where $w$ represents the dimensions of the model and $v$ represents the number of validation samples. A direct comparison for time complexity between \emph{FedVal} and other aggregation methods is difficult, as these other methods mainly depend on number of clients and model dimensions. However, an estimate is that \emph{FedVal} is on the upper middle end of computational load for robust aggregators. In Section \ref{sec:timeexp}, we will further investigate the time complexity by examining the number of validation samples required for \emph{FedVal} and show that \emph{FedVal} only requires a limited number of validation samples to be effective.

In conclusion, the design of \emph{FedVal}  will provide a relatively light-weight and flexible defense that is prepared for any potential scenario, while promoting fairness among labels and continuously striving to reduce the loss on the validation dataset during each training iteration, which can be shown in more detail from our experiments in Section \ref{sec:exp}.
\section{Experimental Setup}

This section will discuss the general setups we have used in our experimental study. Federated learning is simulated with the Virtual Client Engine (VCE) of the comprehensive Flower framework \cite{beutel2020flower} enabling us to scale to a large number of clients within a single machine. Datasets and hyper-parameters are detailed below.

\subsection{Datasets and Partitions}\label{sec:partition}

In our research, we employ three widely used datasets of diverse size and complexity, namely CIFAR-10 \cite{krizhevsky2009learning}, FEMNIST \cite{caldas2018leaf}, and PUMS ACSIncome \cite{ding2021retiring}.

CIFAR-10 is a renowned benchmark dataset in computer vision and machine learning, consisting of 60,000 32x32 color images equally divided among 10 distinct categories. 

On the other hand, FEMNIST is an expanded version of the EMNIST dataset \cite{cohen2017emnist}, partitioned among 3597 authors of the handwritten characters and digits. With over 800,000 28x28 grayscale images, FEMNIST is specifically designed to simulate a realistic federated setting where each author is represented by a client. In the FEMNIST dataset, there is a quite noticeable quantity skew across classes. Specifically, considering the division between numbers, capital letters, and lowercase letters, where lowercase letters and capital letters are underrepresented. Each number has approximately 4 times more elements than the letters, and capital letters are more common than lowercase letters, as reported in \cite{cohen2017emnist}.

Additionally, the PUMS ACSIncome dataset \cite{ding2021retiring} was also utilized. The ACSIncome dataset, based on the Public Use Microdata Sample (PUMS) from the American Community Survey (ACS), provides income-related data on individuals, making it an excellent resource for studying aspects such as demographic bias and fairness in ML applications \cite{ezzeldin2021fairfed, globus2022algorithmic}.

This work has employed several general federated settings for evaluations. The first setting utilized 40 clients for the CIFAR-10, with each client having a dataset of 1250 images. The experiment was conducted over 60 rounds, and 10 clients were selected at random in each round, using a fixed set seed to ensure a consistent selection of data across experiments. This pseudo-random selection of clients ensured that the same data was used for comparison purposes in each experiment.

The second setting is on the FEMNIST experiments, which were performed using the 3597 authors as clients. Each client has a varying number of samples, with an average of around 225 samples per client. The FEMNIST experiment was run over 200 rounds, and in each round, 30 clients were randomly selected in the same pseudo-random fashion as before to ensure consistent data selection for comparisons. For experiments displayed in bar graphs with FEMNIST, average result over five rounds are displayed.

The third setting involves experiments with the ACSIncome dataset. Here, we specifically investigated the recall rate for minority groups as a measure of fairness. We utilized 40 clients for these experiments, which were conducted over 30 rounds with 15 clients selected for each round. Client selection was done in similar pseudo-random fashion as for previous settings and average result over multiple rounds were presented.

In all experimental conditions, we conduct a series of tests, opting to feature the results with the lowest performance for each respective algorithm in our study, unless otherwise stated. Our intent behind this approach is to portray precisely the influence of heterogeneity issues and poisoning attacks on a federated training system. By focusing on these results, we're able to highlight potential instabilities that would remain undetected if we merely displayed averages.

To study the impact of heterogeneity, a range of techniques have been utilized to manipulate client distribution in the CIFAR-10 dataset. We follow the latent Dirichlet distribution (LDA) implemented by \cite{li2022federated, yurochkin2019bayesian,hsu2019measuring} where both label distribution and quantity distribution is determined by input parameter $\alpha$. The level of heterogeneity is governed by the parameter $\alpha$. As $\alpha \to \infty$, partitions become more uniform (IID), and as $\alpha \to 0$, partitions tend to be more heterogeneous. Our experimental evaluation considers both $\alpha=0.4$ and $\alpha=1000$ for the non-IID and IID cases. 

Additionally, this paper introduces a new method to examine fairness in the federated setting. This method artificially creates a situation where a less common type of client contains some vital data for the model. 

On the other hand, in the FEMNIST dataset, heterogeneity occurs naturally due to the unique writing styles of each author, as well as differences in label distribution and sample quantity between authors.

\vspace{-3mm}
\subsection{Model Architecture and Training Details}

This work employs a similar model for both CIFAR-10 and FEMNIST. We implement a widely adopted CNN neural network for the image classification tasks for both datasets. The model used for both datasets is a convolutional neural network with $6$ convolutional layers, and the kernel size used is $3x3$ pixels. For both datasets, the models are trained with SGD, and the number of local client epochs is set to $10$. Both models employ a learning rate of $0.005$.

In addition to employing a similar model for CIFAR-10 and FEMNIST, a different model architecture was utilized for the PUMS ACSIncome dataset. The model consists of a sequential neural network with four dense layers. It was compiled using the Adam optimizer with a learning rate of 0.0001 and trained with the binary cross-entropy loss function.

Also, we set the \emph{FedVal} specific hyper-parameters $s_1$ to be $3$ for all datasets. Since $s_2$ is an adaptive hyper-parameter, it starts from the $3$ and is chosen adaptively in each round as explained in Section \ref{sec:fedval}. As previously mentioned, the summation in Equation \ref{eq:ConScore} can be extended with other dimensions that matter. Our implementation adds overall accumulated average loss to the summation where $s_{1,avg}$ is set to $5$ while $s_2$ is redundant since the bias reducer term will always be $1$ due to dividing overall loss by itself. For all experiments in the paper, the $C$ hyper-parameter is set to $3$. The implemented score function is demonstrated in Equation \ref{eq:ScoreImpl}.

\begin{flalign}\label{eq:ScoreImpl}
    S(\theta_d^t) = &\sum_{k=1}^{K} (\max(1, (\frac{\barfix{\Bar{\pmb{L}}}_{k}}{\pmb{\barfix{\Bar{L}}}_{avg}})^{s_2})*\frac{s_{1,k} *div_{k,d} }{MAD_{k}} + Cs_{1,k}) + &&\\\nonumber
    &\frac{s_{1,avg} *div_{avg,d} }{MAD_{avg}} + Cs_{1,avg}&&
\end{flalign}

In addition to this, in our experiments with ACSIncome, the scoring dimensions were additionally extend to consider recall rate across certain groups. For this dimension the $s_2$ hyperparameter was statically set to $30$.

\subsection{Poisoning Attacks Methods}

We employ two separate poisoning attacks to comprehensively cover the possible types of attacks that might occur in FL systems. The first attack is a \textit{targeted data poisoning attack}, adapted from the method presented by Sun et al. \cite{sun2019can}, which introduces small changes to the model that affect only a limited subset of its tasks. The second attack is an \textit{untargeted model poisoning attack}, adapted from the gradient ascent attack presented by Shejwalkar et al. \cite{shejwalkar2022back}, which seeks to reduce the global accuracy of the model while avoiding detection by the defense mechanisms in place. These types of attacks have been widely studied in the field of poisoning attacks, as documented in recent research works\cite{shejwalkar2022back,shejwalkar2021manipulating,li2019abnormal,karimireddy2020scaffold,sun2019can}. 
 
The underlying principle of the \textit{targeted data poisoning attack} is to manipulate a small subset of the model, making it challenging for defenses to detect the attack. This increases the potential impact on the targeted functionality while minimizing the risk of detection. As a data poisoning attack, it also expands the adversary's reach, allowing for more malicious actors to participate. 
 
On the other hand, the \textit{untargeted model poisoning attack} is designed to cause damage to the model without targeting a specific area of its functionality. With full access to the model during training rounds, the adversary can launch a more powerful attack by finding the optimal malicious direction that the defense will not identify as malicious. This attack is an untargeted \textit{projected gradient ascent attack (PGA)} and aims to impact the model wherever it is feasible.

\subsection{Differential Privacy Methods}

The impact of adding differential privacy on defenses against poisoning attacks in federated learning systems has been the subject of extensive research and discussions \cite{kairouz2021advances, sun2019can}. In this work, we contribute to this body of research by incorporating differential privacy techniques into the federated learning system. We adopt the approach proposed by Andrew et al. \cite{andrew2021differentially} by adaptively binding norms and adding noise to each update vector. 
 
Our aim is to investigate the behavior of various defenses under these conditions and assess the reduction in the impact of many poisoning attacks \cite{sun2019can,panda2022sparsefed}. Our work expands upon previous research by exploring the interplay between differential privacy and defenses against poisoning attacks in federated learning systems. 

Due to the adaptive nature of the approach proposed by Andrew et al., a quantitative evaluation of privacy is not feasible and has been left out of this paper.

\subsection{Baselines and comparison}
We implement various state-of-the-art solutions as baselines from a wide range of prospective in our experiments for comprehensive comparisons.

 
\textit{LFR} \cite{fang2020local} is a poisoning defense that removes a specified number of clients in each round based on their accumulated loss on a server-side validation dataset. In our experiments, $40\%$ of clients were removed each round, as determined by the mathematical experiment in Section \ref{sec:expmal} and the number of malicious clients used. LFR serves as a benchmark algorithm for validation-loss-based defenses. It is worth mentioning that following the original paper, the FEMNIST dataset experiments with LFR use a heterogeneous validation dataset with the same label distribution as the training dataset.
 
\textit{Multi-Krum} \cite{blanchard2017machine}, an outlier-based defense method, is a defense that removes a specified number of client updates deemed as outliers, based on the Euclidean distance between updates. As Multi-Krum (and other outlier-based defenses) struggle to detect model poisoning attacks~\cite{kasyap2022hidden,shejwalkar2022back}, the number of clients to remove for multi-Krum is chosen to be $50\%$.
 
We also used \textit{norm-bounding} \cite{sun2019can} as a benchmark defense to mitigate the effects of poisoning attacks. This defense is integrated into our differential privacy solution \cite{andrew2021differentially} and was evaluated both in combination with other defenses and as a standalone measure against the attacks. We use a target quantile of 50\% of norms to be bound as in \cite{andrew2021differentially}.

Lastly, we implement \textit{FedProx} \cite{li2020federated}. FedProx is a widely-adopted method targeting to solve the data heterogeneity problem by preventing model drift from client updates. FedProx is implemented to investigate shortcomings and potential issues with such solutions. Experiments with FedProx follow the same protocol suggested in the original paper, which tests the hyper-parameter $\mu$ from the range of $[1,5]$.

As mentioned before, fairness and reducing group bias in federated learning is a field not well researched, and existing solutions often use methods that might not be feasible due to privacy and security issues \cite{kairouz2021advances}. This makes it difficult to find baseline solutions targeted at promoting fairness.

\emph{FedVal} distinguishes itself from other existing methodologies in several crucial aspects. While multi-Krum utilizes outlier detection, \emph{FedVal} uncovers malicious clients by analysing accumulated loss on a server-side validation dataset, a strategy similar to that employed by LFR. Despite these similarities, \emph{FedVal} and LFR differ significantly in their operations. LFR relies heavily on pre-existing knowledge of the quantity of malicious nodes, as discussed in Sections \ref{sec:defense1} and \ref{sec:expmal}, can result in complications. In contrast, \emph{FedVal} diverges in its approach by not needing prior knowledge of number of malicious nodes. \emph{FedVal} leverages analyses performed on the server-side validation set to maintain model balance and identify oscillations across specific labels or dimensions. Unlike norm-bounding, which merely aims to minimize the fallout of an attack, \emph{FedVal} seeks to eradicate malicious client updates altogether. Furthermore, for heterogeneity issues, it stands apart from FedProx by striving to craft a model that excels by its very design, rather than simply aligning it with the majority client model.
\vspace{-3mm}
\section{Experiments}
\label{sec:exp}
In this section, we will present the results of our experiments and compare them with the chosen baseline solutions. We will also discuss the significance of the results and provide an extensive analysis of the performance of different algorithms. To begin with, we will present the results of the baselines with different algorithms in Section \ref{sec:baseexp}. We will then move on to the results from poisoning attacks in Section \ref{sec:exppois}. Additionally, we will showcase an analysis of the limitations and potentials of different algorithms in Section \ref{sec:expmal}. Finally, we will demonstrate some results regarding fairness in the non-IID settings in Section \ref{sec:fairexp}.

\subsection{Baselines Results}
\label{sec:baseexp}

\begin{figure}
     \centering
     \begin{subfigure}[b]{0.23\textwidth}
         \centering
         \includegraphics[width=\textwidth]{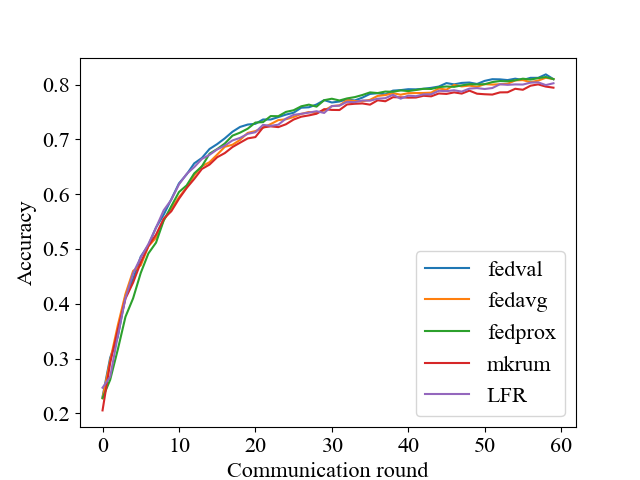}
         \caption{}
         \label{fig:cif}
     \end{subfigure}
     \begin{subfigure}[b]{0.23\textwidth}
         \centering
         \includegraphics[width=\textwidth]{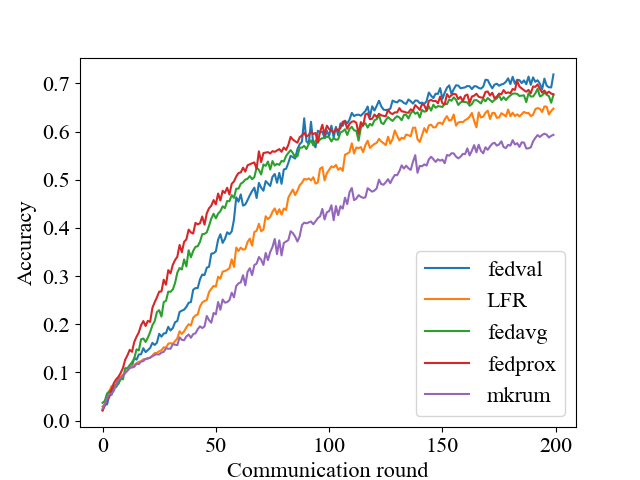}
         \caption{}
         \label{fig:fem}
     \end{subfigure}
\caption{Baseline test accuracy for CIFAR-10 and FEMNIST for different baseline methods. (a) accuracy for CIFAR-10 with IID distribution; (b) accuracy for the naturally heterogeneous FEMNIST dataset.}
\label{fig:baseline}
\end{figure}

Figure \ref{fig:baseline} demonstrates the baseline results for all the algorithms we use. Figure \ref{fig:cif} illustrates the IID test accuracy on CIFAR-10, and the results are what one would expect. Solutions that remove a set number of clients each round, such as LFR and multi-Krum have a very slight decrease in accuracy, while all other methods perform similarly. However, the results on the naturally heterogeneous dataset FEMNIST show more diverse results. All robust aggregators converge a bit slower than FedAvg and FedProx, which are methods not targeting for defense, but we can see that \emph{FedVal} converges to higher accuracy than other solutions, while multi-Krum and LFR converge to lower accuracy. It is worth noting that the results from FEMNIST are from IID test data, which means that, as opposed to the train data, the test data has the same amount of elements for each label. 

\subsection{Poisoning attacks}
\label{sec:exppois}
This section aims to evaluate the resilience of \emph{FedVal} against other baseline algorithms with the presence of various forms of poisoning attacks in diverse settings, including systems with differential privacy and different types of heterogeneity.

\begin{figure}
     \centering
     \begin{subfigure}[b]{0.23\textwidth}
         \centering
         \includegraphics[width=\textwidth]{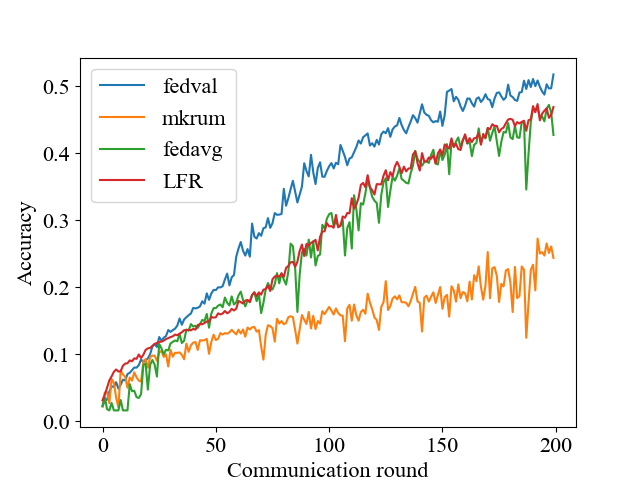}
         \caption{}
         \label{fig:fpga1}
     \end{subfigure}
     \begin{subfigure}[b]{0.23\textwidth}
         \centering
         \includegraphics[width=\textwidth]{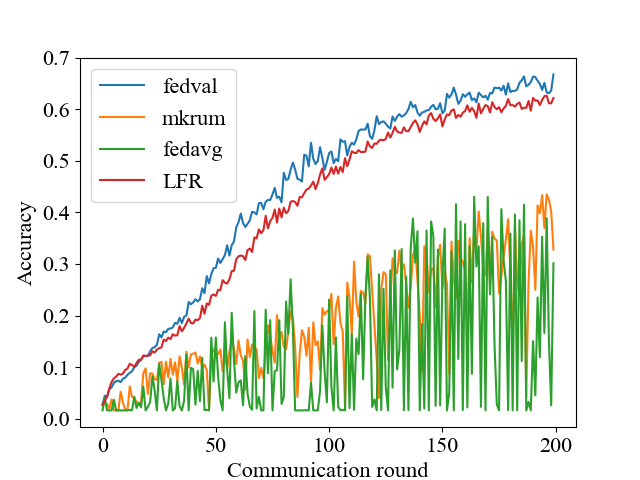}
         \caption{}
         \label{fig:fpga2}
     \end{subfigure}
\caption{Comparison of accuracy for various aggregation methods applied to the naturally heterogeneous FEMNIST dataset under PGA attack conditions, with and without the application of the differential privacy techniques of binding norms and noise addition. (a) PGA attack scenario with differential privacy techniques, (b) PGA attack scenario without differential privacy techniques.}
\label{fig:femnist}
\end{figure}

To begin with, in Figure \ref{fig:femnist}, we illustrate the accuracy of different baseline algorithms when the system is under a static PGA model poisoning attack by $10\%$ of the present clients in the FEMNIST dataset. In our experiments, we have considered two different scenarios - one where noise is added and norms are cut by the value approximating $50\%$ of norms being cut, illustrated in Figure \ref{fig:fpga1}, and another where there is no noise or norm-bounding, illustrated in Figure \ref{fig:fpga2}. 

Results demonstrate that \emph{FedVal} performs the best in both cases. This can be attributed to its natural ability to handle heterogeneous data and the ability not to ignore any good client models due to any set limit on the potential number of malicious clients. The results indicate that both \emph{FedVal} and LFR are able to remove all malicious updates that would have any noticeable impact, but LFR differs in that it also removes a significant number of benign updates, since 40\% of clients in an LFR update are always removed. Interestingly, in the norm-bounding scenario depicted in Figure \ref{fig:fpga1}, multi-Krum performs worse than simply averaging (FedAvg). This is likely due to the fact that the norms of the malicious updates are scaled in such a way that they are not deemed malicious by the aggregator. As a result, since half of all clients are removed in our multi-Krum implementation, this leads to a larger percentage of remaining clients being malicious. 

\begin{figure}
     \centering
     \begin{subfigure}[b]{0.23\textwidth}
         \centering
         \includegraphics[width=\textwidth]{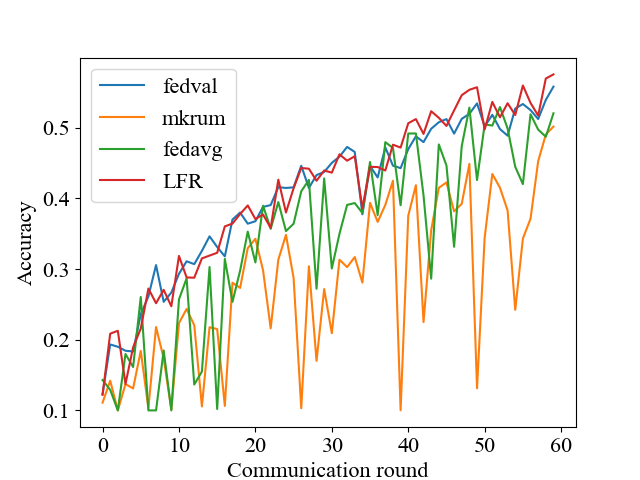}
         \caption{}
         \label{fig:cdp1}
     \end{subfigure}
     \begin{subfigure}[b]{0.23\textwidth}
         \centering
         \includegraphics[width=\textwidth]{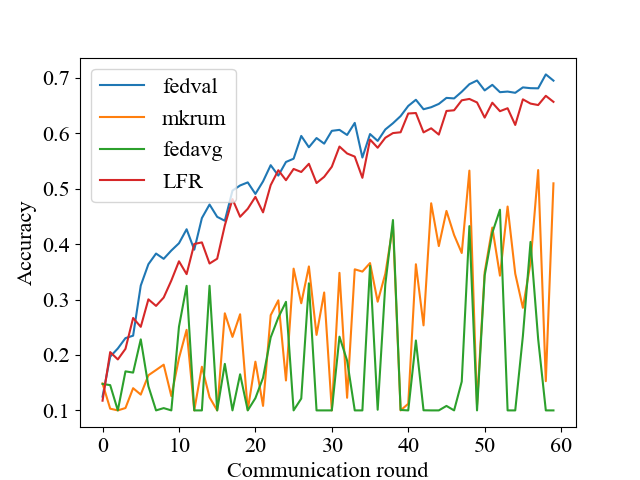}
         \caption{}
         \label{fig:cdp2}
     \end{subfigure}
\caption{Comparison of accuracy for various aggregation methods applied to the CIFAR-10 dataset, targeted under PGA attack conditions, with simulated heterogeneity in client data distribution using LDA $(\alpha=0.4)$. (a) PGA attack scenario with differential privacy techniques (binding norms and noise addition), (b) PGA attack scenario without differential privacy techniques.}
\label{fig:cifar}
\end{figure}

In addition, in Figure \ref{fig:cifar}, we present results from experiments that are similar to those shown in Figure \ref{fig:femnist}, using the CIFAR-10 dataset. In this experiment, we have introduced artificial heterogeneity by utilizing the LDA with $\alpha=0.4$ as explained in Section \ref{sec:partition}. Similar to the previous experiment, we compare different baseline algorithms and \emph{FedVal} under the condition where $10\%$ of present clients are malicious, with the noise and norm-bounding setting on Figure \ref{fig:cdp1}, and without noise and norm-bounding Figure \ref{fig:cdp2}. From these experiments, we can see that \emph{FedVal} and LFR perform similarly under the noise and norm-bounding conditions, while \emph{FedVal} slightly outperforms LFR without noise and binding norms.

It is important to note that \emph{FedVal} is designed with fairness across labels as a goal, with the aim of improving the accuracy of underrepresented labels. This feature increases the accuracy of the test dataset performed with the FEMNIST dataset since there are underrepresented labels present, such that some labels are present in less quantity. However, with the use of the LDA on CIFAR-10, the underlying data distribution remains homogeneous, but the client data distribution is skewed. This is likely the reason for the difference in results between Figure \ref{fig:femnist} and \ref{fig:cifar}. On the other hand, we can observe the same trends for CIFAR-10 in  Figure \ref{fig:cifar} as we did for FEMNIST in Figure \ref{fig:femnist}. The results presented in both figures demonstrate the effectiveness of \emph{FedVal} in improving the performance of FL under various conditions, including label distribution skew and with the presence of poisoning attacks. 

\begin{figure}
     \centering
     \begin{subfigure}[b]{0.23\textwidth}
         \centering
         \includegraphics[width=\textwidth]{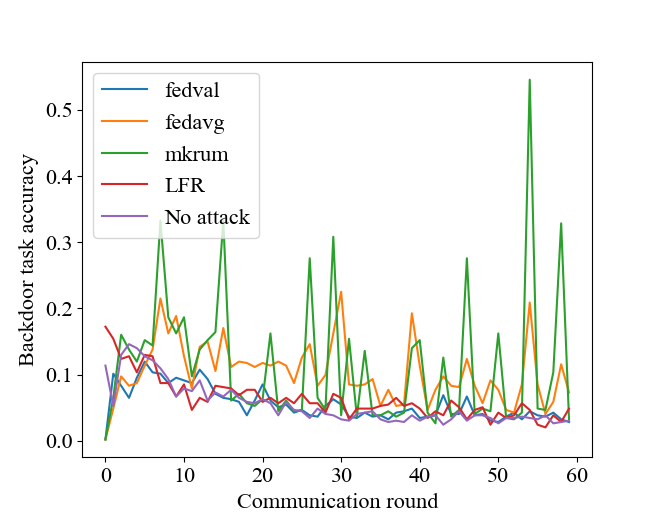}
         \caption{}
         \label{fig:horse1}
     \end{subfigure}
     \begin{subfigure}[b]{0.24\textwidth}
         \centering
         \includegraphics[width=\textwidth]{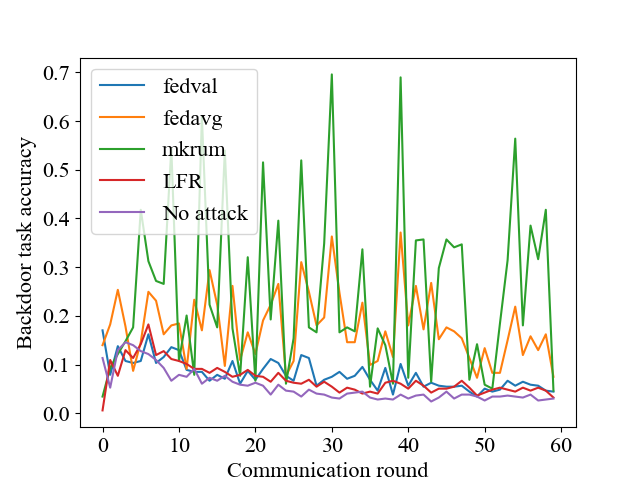}
         \caption{}
         \label{fig:horse2}
     \end{subfigure}
\caption{Backdoor performance over rounds for the backdoor task of miss-predicting horses into being deer. Malicious clients perform data poisoning attacks by swapping the labels of `horses' into `deer'. The backdoor accuracy is measured by how often the model incorrectly predicts a `horse' image as a `deer'. (a) attack accuracy with 10\% malicious clients; (b) attack accuracy with 20\% malicious clients.}
\label{fig:backdoor}
\end{figure}

Furthermore, we conduct experiments with backdoor poisoning attacks on the CIFAR-10 dataset as shown in Figure \ref{fig:backdoor}. In the experiment, we manipulate the label of the data by changing the labels of `horse' images to `deer'. Specifically, in Figure \ref{fig:horse1}, we change $10\%$ of the clients' data, and in Figure \ref{fig:horse2}, we change $20\%$ of the clients' data. The accuracy of the backdoor attack in this experiment is measured by how often the model incorrectly predicts a `horse' image as a `deer', so the lower accuracy indicates less successful attacks. By investigating the results, we can see that the LFR and \emph{FedVal} aggregation methods outperform other methods in protecting the model against the backdoor attack. Also, the results show that both LFR and \emph{FedVal} are able to provide almost full protection for the model, as compared with the no-attack case. This highlights the effectiveness of our method in mitigating the impact of malicious actors attempting to manipulate machine learning models using backdoor attacks.

\vspace{-3mm}
\subsection{Number of malicious clients}
\label{sec:expmal}

To ensure robustness, it is important to consider the potential presence of a percentage of malicious clients in a round over a large number of rounds. More details regarding the theoretical scenarios are explained in Appendix \ref{app:numclients}. Following the discussion in Section \ref{sec:defense1} and Appendix \ref{app:numclients}, we decided to investigate the situation where the number of malicious clients exceeds the limit set by robust aggregators in order to understand the capabilities of our method \emph{FedVal} fully. For other aggregators, as one might expect, the model's performance deteriorates significantly in scenarios where the number of malicious clients exceeds the limit set by a robust aggregator.

\begin{figure}
     \centering
     \begin{subfigure}[b]{0.23\textwidth}
         \centering
         \includegraphics[width=\textwidth]{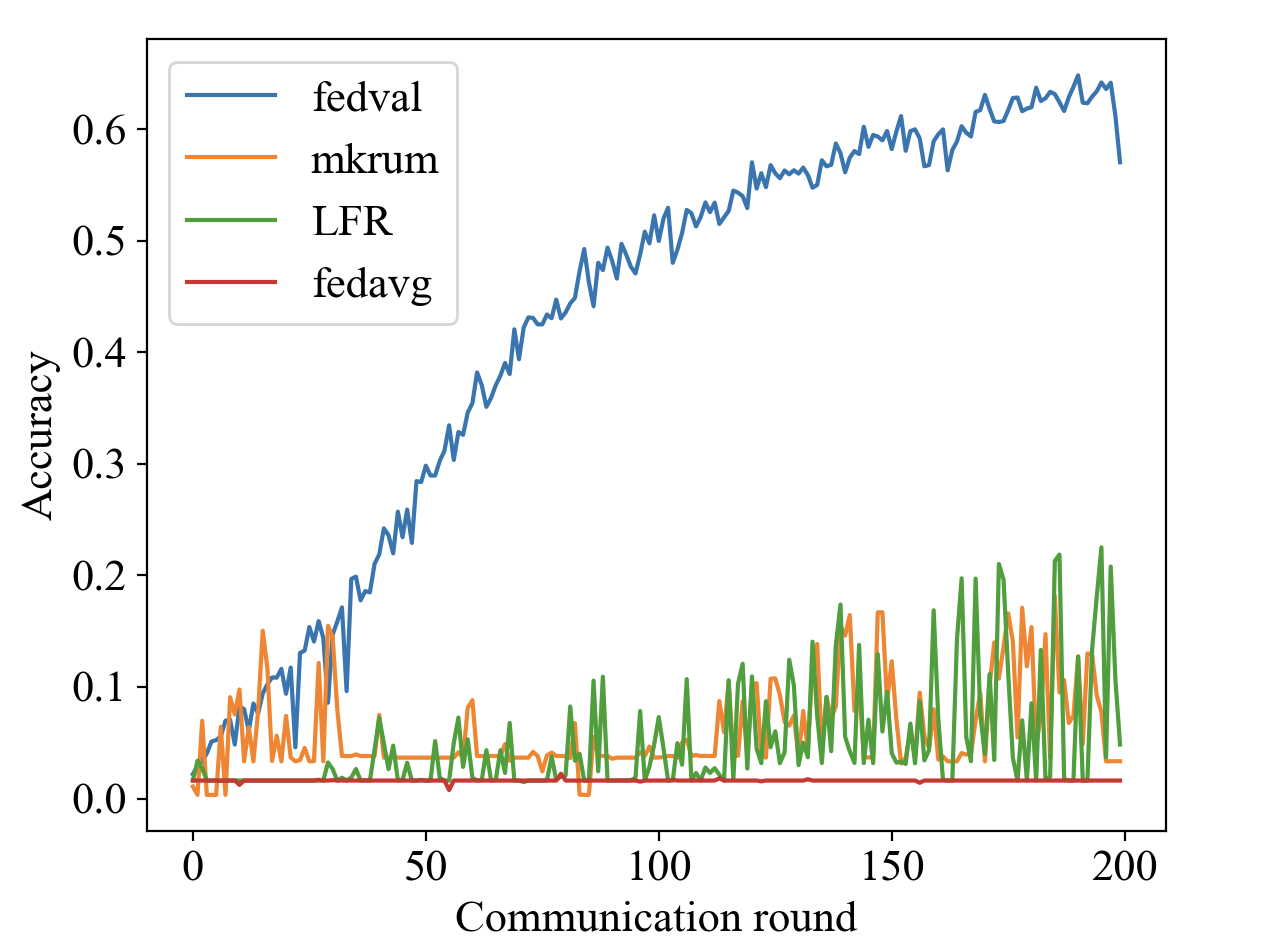}
         \caption{}
         \label{fig:no1}
     \end{subfigure}
     \begin{subfigure}[b]{0.23\textwidth}
         \centering
         \includegraphics[width=\textwidth]{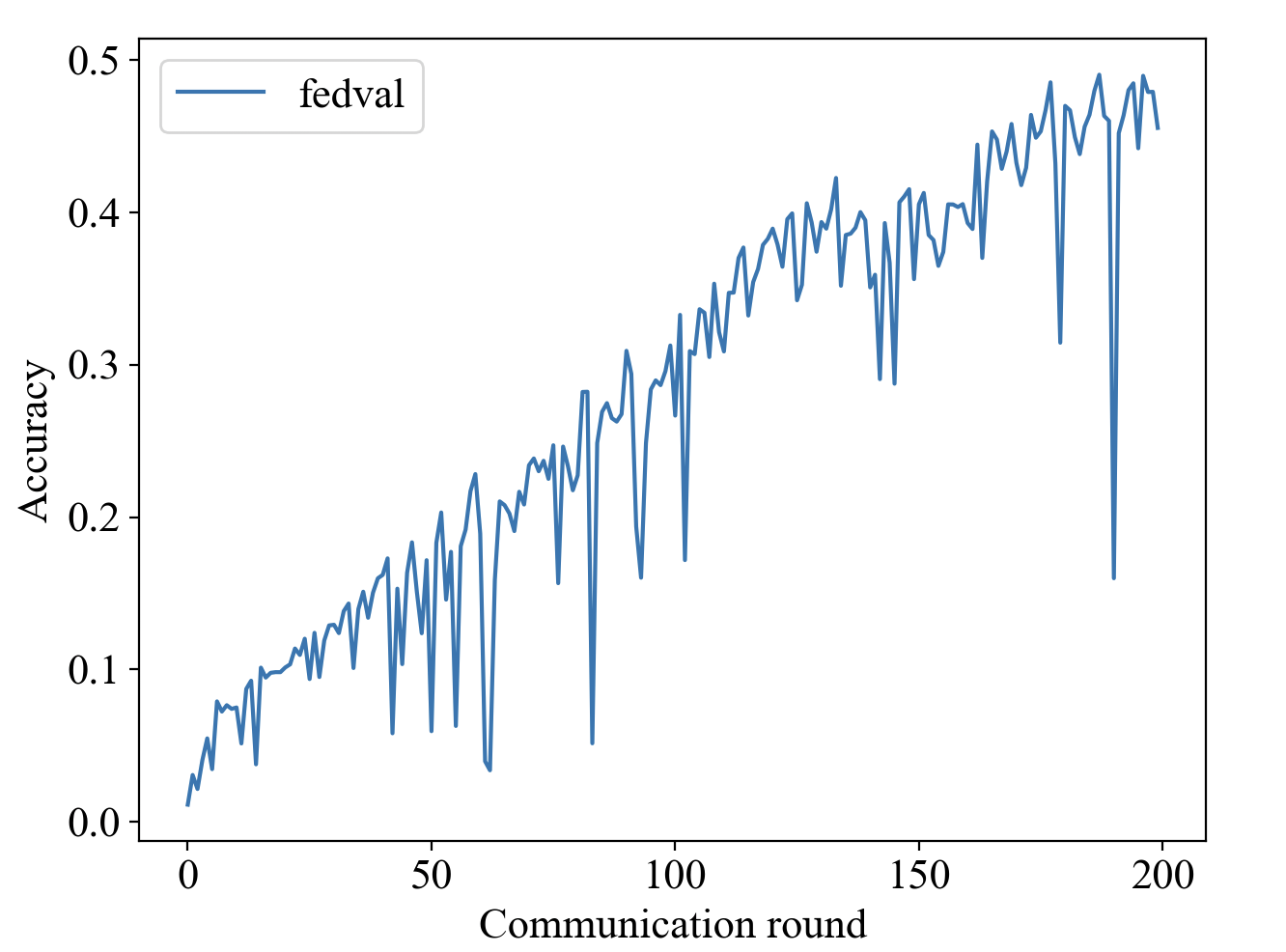}
         \caption{}
         \label{fig:no2}
     \end{subfigure}
\caption{Performance of various defenses under stress test conditions with a large number of malicious clients conducting poisoning attacks. (a) accuracy with 40\% of clients being malicious, (b) accuracy with 80\% of clients being malicious.}
\label{fig:femnistnumber}
\end{figure}

We conducted experiments with both $40\%$ and $80\%$ of total clients as malicious, illustrated in Figure \ref{fig:femnistnumber}, and the results were quite striking. Despite the high percentage of malicious clients, \emph{FedVal} was able to converge without any major difficulties as opposed to other methods. This highlights the strength of \emph{FedVal} as both a robust but also adaptive poisoning defense that is prepared for any type of situation that may occur in a federated learning system.

Furthermore, the results of these experiments also demonstrate the robustness of \emph{FedVal} in the face of a high number of malicious clients. In the standard federated learning system using aggregation methods such as FedAvg, the presence of even a small number of malicious clients can cause serious issues with the convergence and accuracy of the model, as we can see from Figures \ref{fig:femnist}, \ref{fig:cifar}, and \ref{fig:backdoor}. However, our method can mitigate the impact of malicious clients and ensure the integrity of the model in various scenarios.

Overall, the experimentation in Figure \ref{fig:femnistnumber} has shown that \emph{FedVal} is an effective and reliable method for defending against poisoning attacks in federated learning systems, even in situations where the number of malicious clients is higher than what we may expect there to be.

\vspace{-3mm}
\subsection{Time complexity}\label{sec:timeexp}

\begin{figure}[t!]
     \centering
     \begin{subfigure}[b]{0.23\textwidth}
         \centering
         \includegraphics[width=\textwidth]{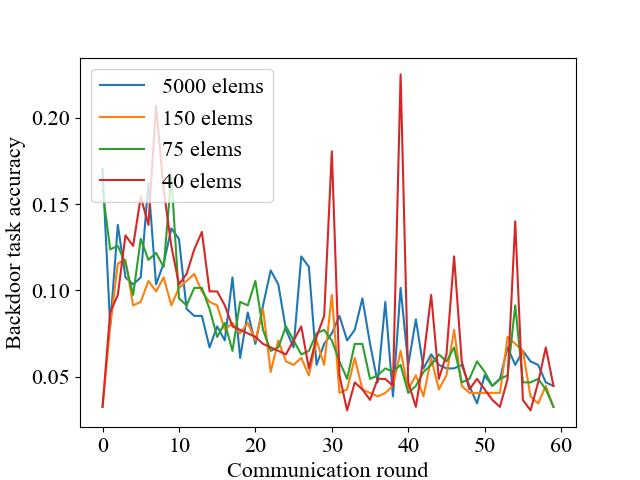}
         \caption{}
         \label{fig:val1}
     \end{subfigure}
     \begin{subfigure}[b]{0.23\textwidth}
         \centering
         \includegraphics[width=\textwidth]{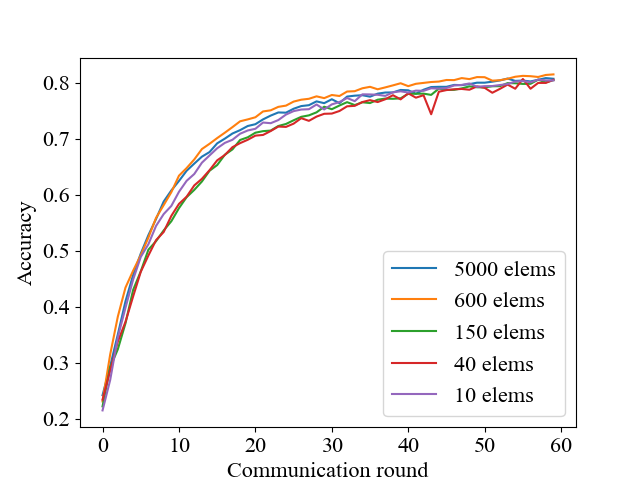}
         \caption{}
         \label{fig:val2}
     \end{subfigure}
\caption{Performance of \emph{FedVal} under poisoning attacks with different numbers of validation elements used for the analysis on CIFAR-10. (a) accuracy for the backdoor task of the malicious clients, (b) overall accuracy.}
\label{fig:time}
\end{figure}

We discussed the time complexity and computational load of running \emph{FedVal} previously in Section \ref{sec:timeexp}. To further investigate this issue, we have conducted an experimental study to determine the minimum number of validation elements required for the algorithm to maintain robustness. We present the experiment results in Figure \ref{fig:time}, which are conducted using the CIFAR-10 dataset and various attacks. We varied the number of validation elements used in the defense and evaluated the effectiveness of the system under these conditions.

The experiment shows that a relatively small number of validation elements is sufficient to ensure robustness. For the untargeted attack illustrated in Figure \ref{fig:val2}, we obtained the unexpected result that checking the loss on a single element for each class is enough to achieve robustness. For the targeted backdoor attack, illustrated in Figure \ref{fig:val1}, $75$ total validation elements are required to achieve robustness, which is $7.5$ elements per class.

Based on these findings, it can be inferred that using $10$ validation elements per class provides a considerable degree of robustness, given the experimental conditions we've explored. We recommend maintaining a validation dataset that is minimalistic yet comprehensive in scope. This provides important information, as it suggests that the \emph{FedVal} algorithm is computationally efficient and can be implemented with minimal computational resources.


\begin{figure}[t!]
\centering
\includegraphics[width=0.5\linewidth, trim=1cm 0cm 0cm 0cm]{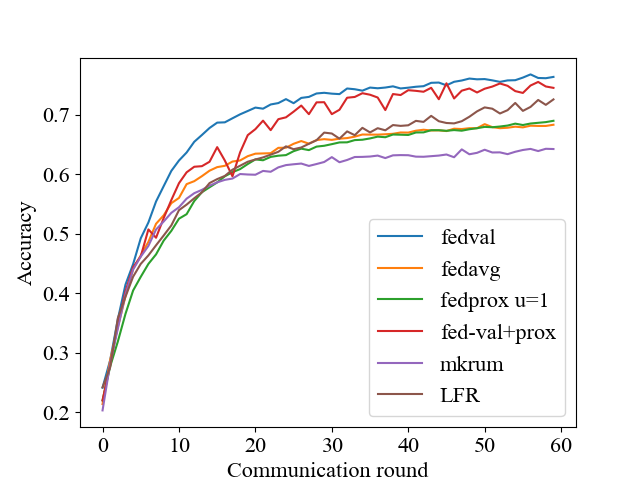}
\caption{Test on CIFAR-10 non-IID. Two labels are missing in a majority of the clients. The clients who have the missing labels are IID.}
\label{fig:missing}
\end{figure}

\vspace{-3mm}
\subsection{Fairness} \label{sec:fairexp}

\begin{figure}[t!]
     \centering
     \begin{subfigure}[b]{0.28\textwidth}
         \centering
         \includegraphics[width=\textwidth]{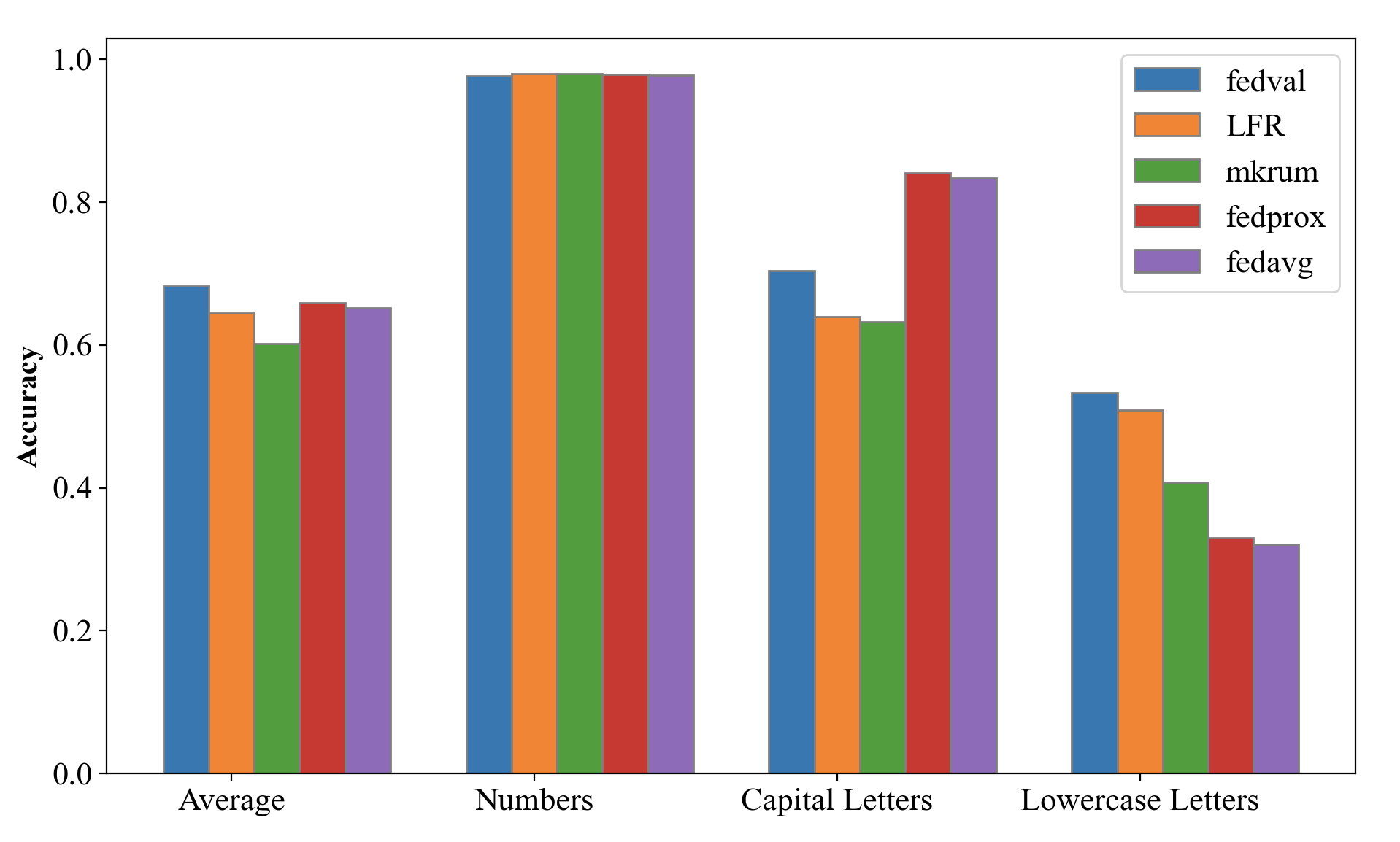}
         \caption{}
         \label{fig:fair1}
     \end{subfigure}
\hfill
     \begin{subfigure}[b]{0.28\textwidth}
         \centering
         \includegraphics[width=\textwidth]{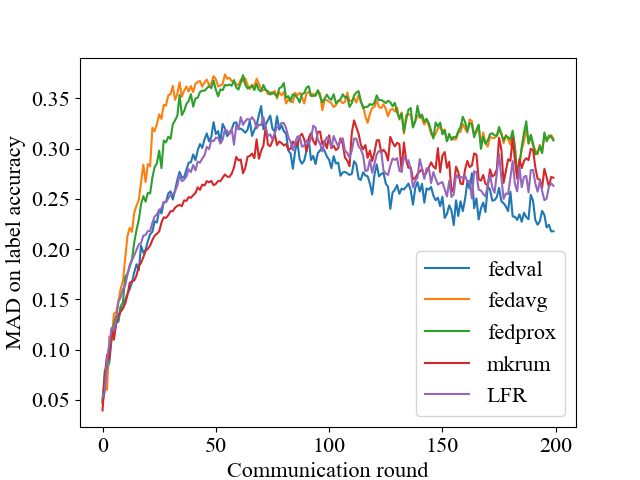}
         \caption{}
         \label{fig:fair2}
     \end{subfigure}
\hfill
     \begin{subfigure}[b]{0.28\textwidth}
         \centering
         \includegraphics[width=\textwidth]{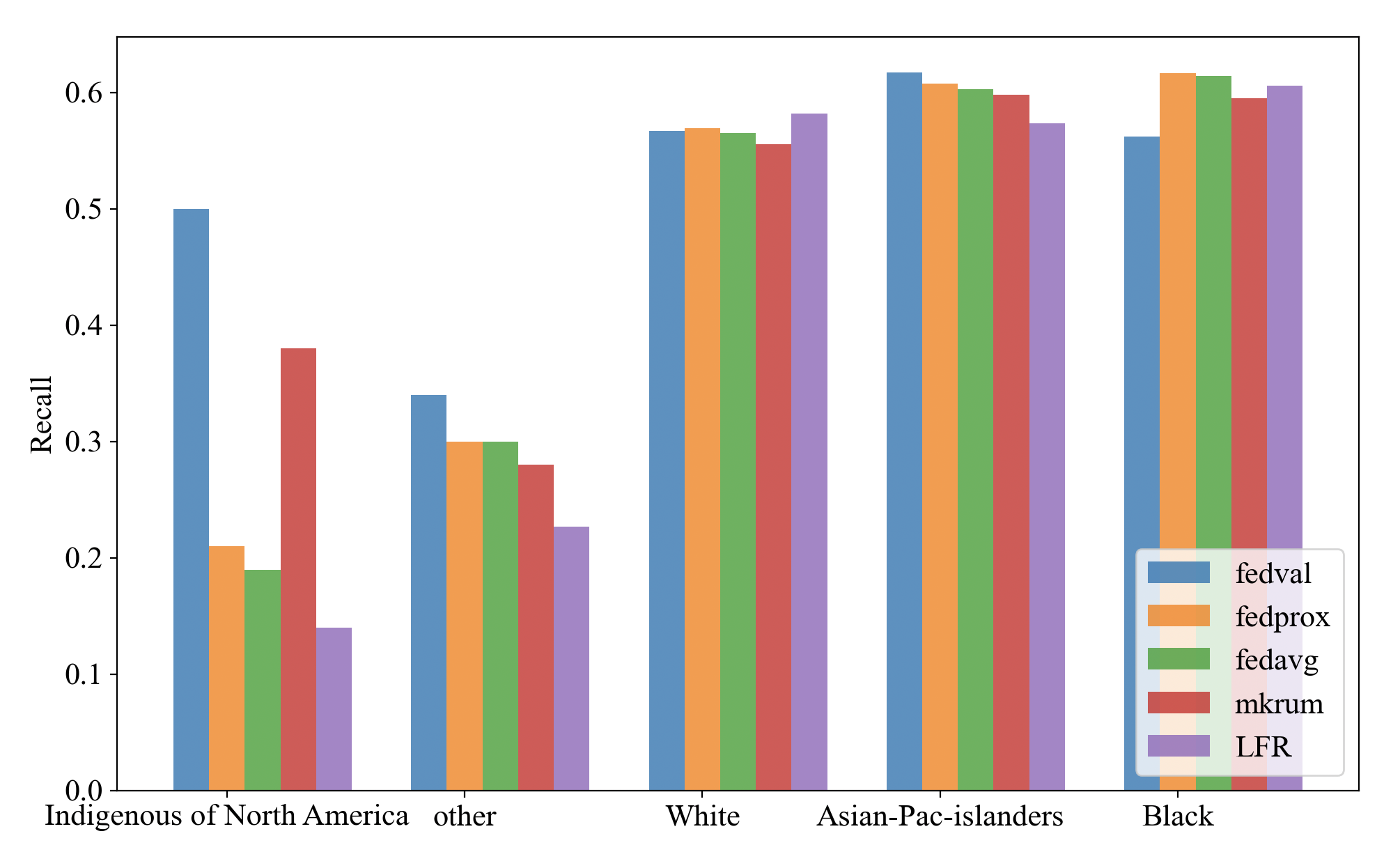}
         \caption{}
         \label{fig:fair3}
     \end{subfigure}
\caption{Fairness experiment on the naturally heterogenous FEMNIST and PUMS ACSIncoms dataset. (a) average accuracy for the final 5 rounds on each class group on FEMNIST, (b) mean absolute deviation across all labels on FEMNIST, (c) average recall for final 10 rounds on different groups in ACSIncome.}
\label{fig:fair}
\end{figure}
Continuing our experimental study, we will now focus on investigating the impact of heterogeneous data in federated learning. Specifically, we will demonstrate the importance of handling fairness in federated learning, especially in scenarios  where certain classes are underrepresented in some clients.

To illustrate this point, we have designed a simple scenario using the CIFAR-10 dataset, where $2$ of the $10$ classes are missing in the $70\%$ of the participating clients. Figure \ref{fig:missing} presents the results of this scenario, where classes `4' and `5' are the missing classes. The results indicate that in this scenario, the accuracy for classes `4' and `5' hover around zero for a majority of the tested solutions. Solutions managing to handle the scenario are our proposed solution, \emph{FedVal} and interestingly also to a certain degree with LFR. This is likely attributed to the fact that LFR would favor the clients with the complete label distribution and remove a few skewed clients each round. This is not sufficient to handle the setting as the clients with the missing labels aggregation needs to be scaled up for the missing labels to be properly addressed, as we can see happening with \emph{FedVal}.

We have also implemented FedProx for this scenario, which is a widely-adopted solution for non-IID data distribution in federated learning to prevent client drift. As mentioned, preventing client drift is beneficial in situations where the optimized global model is close to the type of model that a majority of the clients would have. However, in scenarios where some labels are only present in a small subset of all clients, solutions like FedProx falls short, as we can see from the results presented in Figure \ref{fig:missing}. The results where FedProx is evaluated in combination with \emph{FedVal} even indicate that FedProx increases the difficulty of extracting the information that is missing in the majority of clients, likely due to the penalty term, which prevents the clients from moving the model too much, resulting in all clients having more similar models as the ones with missing labels. On the other hand, by utilizing re-weighting schemes, \emph{FedVal} is able to salvage this.

These results demonstrate the importance and significance of fairness in the federated learning system, especially when many clients have underrepresented or missing classes. The results also highlight the catastrophic forgetting in the federated setting \cite{mccloskey1989catastrophic, huang2022learn}, which can be a crucial issue to consider, especially in practical and real-life scenarios, where some labels likely will be less common and not present in all clients. 


Relating to the experiment in Figure \ref{fig:missing}, we also investigate the accuracy of the model thoroughly across all labels in the FEMNIST dataset. As mentioned in Section \ref{sec:partition}, there is a quite noticeable quantity skew across classes. Specifically, considering the division between numbers, capital letters, and lowercase letters, each class of Numbers has approximately 4 times more elements than the Letters, and Capital Letters are more common than Lowercase Letters. With this type of distribution, we can expect the model to quickly learn how to correctly classify Numbers, and struggle with correctly classifying capital and lowercase letters in comparison.

In Figure \ref{fig:fair}, we illustrate the performance of different aggregation methods on different label classes. Figure \ref{fig:fair1} illustrates the accuracy division across label groups, and Figure \ref{fig:fair2} illustrates the mean absolute deviation across all labels. From Figure \ref{fig:fair1}, we can see that by simply averaging through FedAvg or using methods that prevent client drift, such as FedProx, a large bias is created toward the more common groups. The models created by these methods tend to bias toward the more common groups (the Numbers group in our case) and neglect the less common groups, which are the Lowercase Letters. However, most of the robust aggregators happen to create less bias in the model. The most noticeable difference is with \emph{FedVal}, which manages to almost double the accuracy of the less common Lowercase Letters, more exact from $32\%$ with FedAvg to $53\%$ with \emph{FedVal}, while still managing to provide similar accuracy for the more common classes. On the other hand, other methods struggle with creating a model that predicts the Lowercase Letters correctly.

In Figure \ref{fig:fair3}, we have conducted an analysis utilizing the PUMS ACSIncome dataset \cite{ding2021retiring}. This particular dataset is frequently employed to examine fairness issues \cite{ezzeldin2021fairfed, globus2022algorithmic}. The focus of our study was to scrutinize the recall rate (true positive) for minority groups. Historically, ML models have demonstrated a bias towards groups with fewer data points, resulting in a lower rate of accurate positive predictions. 

In our experiment, we extended the dimensions of \emph{FedVal} to encompass discrepancies in recall rates, aiming to equalize recall across all classes. This experiment's findings underscore \emph{FedVal's} dynamic capacity to adjust a model where necessary. By extending the algorithm to include recall analysis, we observed a significant improvement in recall rates between the "Indigenous of North America" and "Other" groups. Notably, these are groups where other methods have previously struggled to generate accurate positive predictions based on true labels.

In conclusion, to improve the accuracy of the model across all labels and ensuring we have a balanced model, we can consider using more advanced aggregation methods such as \emph{FedVal}. Overall, it is important to consider the dataset distribution and potential biases when training and evaluating FL models to ensure that the model is accurate and fair across all classes and demographic groups.

\vspace{-3mm}
\subsection{Summary} \label{sec:sum}

Many existing solutions tend to overlook other critical problems in Federated Learning, often exacerbating these issues or introducing new security and privacy challenges. In contrast, our proposed method, \emph{FedVal}, has shown resilience against both model and data poisoning attacks. It even thrives when a substantial majority of clients are malicious. For example, \emph{FedVal} manages to converge even with $80\%$ of clients performing model poisoning attacks. This is a significant improvement over methods like multi-Krum and LFR, which fail to converge when faced with merely $40\%$ malicious clients. Furthermore, when faced with $20\%$ of clients deploying backdoor data poisoning attacks, both \emph{FedVal} and LFR maintain their performance levels. In the same scenario, multi-Krum's performance drops drastically, even falling behind the no-defense FedAvg aggregator. In addressing data distribution skews, \emph{FedVal} outshines other methods, salvaging problematic situations where others struggle, as illustrated in the CIFAR-10 experiment in Figure \ref{fig:missing}. Moreover, our fairness experiment using the ACSIncome dataset showed that \emph{FedVal} significantly improves the recall of underrepresented groups from $19\%$ with FedAvg to $50\%$, an increase of over $30\%$. FedProx, which is specifically designed to handle heterogeneity, performed similarly to FedAvg's simple averaging. Other robust aggregators showed some promise, but still fell short in performance compared to \emph{FedVal}.

In conclusion, \emph{FedVal} consistently outperforms existing methods in terms of robustness to attacks, promoting fairness, and adapting to data distribution skews. Our work highlights the necessity of considering multiple aspects of Federated Learning, from security to fairness, to ensure the robustness and reliability of the learning system. 

\section{Conclusion}

In this work, we provide a robust solution, \emph{FedVal}, which aims to solve multiple problems by analyzing and utilizing the clients' learning. 
We propose to do this by using a small server-side validation dataset to asses client updates and determine the optimal aggregation weights considering both robustness and fairness. This technique involves comparing average client performance with client performance over a range of dimensions.


The preceding sections of this paper have underscored the potential of the \emph{FedVal} algorithm in the realm of FL. However, we acknowledge the necessity of extending our investigation to assess its applicability on more general regression problems. The breadth of unexplored challenges within FL suggests a wealth of potential benefits from utilizing client analysis through server-side validation data. It is therefore interesting to delve deeper into these areas, as they could provide strong justification for the broader adoption of \emph{FedVal} and similar algorithms.

Moreover, we also perceive the potential of exploring more sophisticated aggregation methods, exploiting the data acquired through validation. The development of such methods could push the boundaries of what is currently achievable, thereby offering a new dimension of innovation. As we strive for continual improvement in FL, our future research endeavors will focus on unearthing and maximizing the potential inherent in these investigative avenues.



\section*{Availability}

Code available at: https://github.com/viktorvaladi/FedVal


\printbibliography
\appendix
\section{Number of malicious clients}
\label{app:numclients}

\begin{figure}
\centering
\includegraphics[width=0.90\linewidth, trim=1cm 0cm 0cm 0cm]{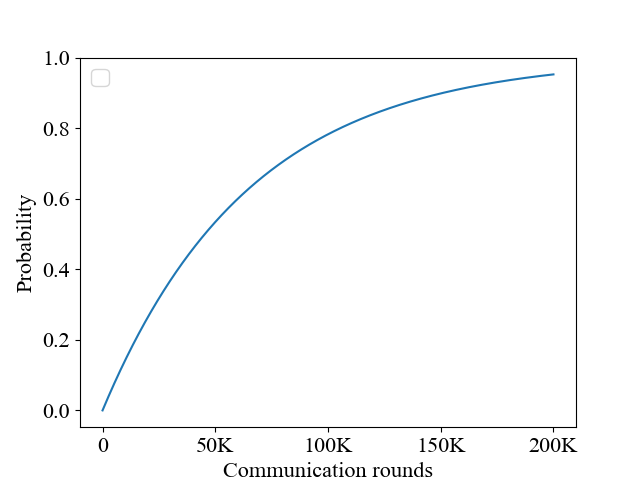}
\caption{Probability to exceed 40\% malicious clients selected in a round at least once when there are 10\% total malicious clients in the system.}
\label{fig:number}
\end{figure}

As mentioned in Section \label{sec:expmal}, in order to ensure robustness, it is important to consider the potential presence of a percentage of malicious clients in a round over a large number of rounds. Figure \ref{fig:number} demonstrates a theoretical scenario in which the number of malicious clients present in the system is $10\%$ of the total clients. To counteract this, we have theoretically implemented protection measures that guarantee robustness when up to $40\%$ of the clients selected for a round are malicious. The figure shows the probability that in at least one round, there will be more than $40\%$ of malicious clients out of selected clients when $30$ clients are selected in each round. The probability was calculated using the binomial distribution $B~(n,k,p)$ with values $n = 30, k = x, p = 0.1$. To get the probability of the event that $9$ or more malicious clients are present, we sum the probabilities of $k >= 9$ up to 30 
\begin{equation}
   P(X) = \sum_{k=9}^{30} B~(30,k,0.1) 
\end{equation}

To get the probability that this event would happen at least once in $x$ number of rounds, we inverse the probability of the event not happening for $x$ consecutive rounds 
\begin{equation}
    1-(1-P(X))^x
\end{equation}

which is the plot given in Figure \ref{fig:number} over communication rounds $x$. As we can see, after $25,000$ rounds, we are very likely to have a "successful attack" where the bound is broken. 

\section{Algorithm}
\label{app:algo}
\begin{algorithm}
\caption{FedVal: $N$ is total number of clients. $r$ is the ratio of selecting clients each round. $D_d$ is the set of client data in client $d$. $\theta_g^t$ is the global model parameter at round t and $\theta_d^t$ is the local model parameter at client $d$. $L$ be the loss function and $\eta$ be the learning rate.} \label{algo:fedval}
\begin{algorithmic}[1]
\Procedure{FedVal}{$s1, s2, ValidationData$}
    \State Initialize model $\theta^0_g$
    \For{$t = 0,1,..,T-1$}
        \State Randomly select set of $rN$ devices $\{S_t\}$
        \State $\pmb{\theta^t_d}$ = \Call{LocalTraining}{$\theta^t_g, S_t$}
        \State $\theta^{t+1}_g, s_2$ = \Call{ScoreClients}{$\{ \theta^t_d: d \in S_t \}, s1, s2$}
    \EndFor

\Function{LocalTraining}{$\theta^t_g, S_t$}
    \For{$d \in S_t$ in parallel}
        \State Update model with client data $D_d$ 
        \State $\theta_d^t \leftarrow \theta^t_g - \eta \nabla L( \theta^t_g; D_d )$ 
    \EndFor
    \State \Return $ \{ \theta^t_d: d \in S_t \} $
\EndFunction
\newline

\Function{ScoreClients}{$\{ \theta^t_d: d \in S_t \}, s1, s2$}
    \For{$d \in S_t$}
        \State Calculate losses $\pmb{L_d} = L(\theta^t_d, validationData)$
        \State Calculate diversions from mean per dimension 
        \State $\pmb{div}_{k,d} = \Bar{\pmb{L}}_{K} - \pmb{L}_{k,d}$, for $k = 1,...,K$
    \EndFor
    \State Calculate mean absolute deviation 
    \State $\pmb{MAD}_{k} = \frac{\sum_{m=1}^{M}{| \pmb{L}_{m,k} -\sum_{i=1}^{M} \pmb{L}_{i,k}|}}{rN}$
    \For{$s_2, s_2+0.5, s2-0.5, s_2+5, s_2-5$ in parallel}
        \For{$d \in S_t$}
            \State $S(\theta_d^t) =$ Equation \ref{eq:ConScore}
        \EndFor
        \State Aggregate by current $s_2$ value 
        \State $\theta^{t+1}_{s2} = \theta^{t+1}_g + \sum_d{\frac{max(0,S(\theta_d^t)*\theta^t_d)}{\sum_d{max(0,S(\theta_d^t)})}}$
        \State Evaluate loss for $s_2$ by loss function $\mathcal{L}$
        \State $ \pmb{L}_{\theta^t_{s2}} = \mathcal{L}(\theta^t_{s2}, ValidationData)$
    \EndFor
    \State Choose $s_2$ and model with min validation loss
    \State $\theta^{t+1}_g, s_2 \sim min(\pmb{L}_{\theta^t_{s2}})$
    \State \Return $\theta^{t+1}_g, s_2$ 
\EndFunction
\EndProcedure
\end{algorithmic}
\label{alg:FedVal}
\end{algorithm}

\end{document}